\documentclass[journal]{IEEEtran}

\usepackage[normalem]{ulem} 
\usepackage[utf8]{inputenc}
\usepackage[T1]{fontenc}
\usepackage{siunitx}
\usepackage[colorlinks=true, allcolors=blue]{hyperref}
\usepackage{amsmath}
\usepackage{amsfonts}
\usepackage{comment}
\usepackage{tikz}
\usetikzlibrary{shapes.multipart}

\usepackage{xurl}

\usepackage{xcolor}
\definecolor{lime}{HTML}{A6CE39}
\DeclareRobustCommand{\orcidicon}
{
    \begin{tikzpicture}
    \draw[lime, fill=lime] (0,0) circle [radius=0.16] 
    node[white] {{\fontfamily{qag}\selectfont \tiny ID}};    \draw[white, fill=white] (-0.0625,0.095) circle [radius=0.007];    
    \end{tikzpicture}
    \hspace{0mm}}
\foreach \x in {A, ..., Z}{%
    \expandafter\xdef\csname orcid\x\endcsname{\noexpand\href{https://orcid.org/\csname orcidauthor\x\endcsname}{\noexpand\orcidicon}}
}

\usepackage{booktabs}
\usepackage{makecell}
\usepackage{soul} 
\usepackage{multirow}
\usepackage{xspace}
\usepackage{color}
\usepackage{colortbl}
\usepackage{adjustbox}
\usepackage{floatrow}
\usepackage{subcaption}
\usepackage{graphicx}
\usepackage{soul}
\providecommand{\lan}[1]{\textcolor{blue}{lan: {#1}}}
\definecolor{myblue}{RGB}{25, 149, 179}

\begin{document}

\setcounter{page}{253}

\title{Vision-based Human Pose Estimation via Deep Learning: A Survey}

\author{Gongjin Lan$^*$\hspace{-1mm}\orcidA{}\hspace{-1mm}, \IEEEmembership{Member IEEE}, Yu Wu, Fei Hu, \IEEEmembership{Member IEEE}, Qi Hao$^*$\hspace{-1mm}\orcidB{}\hspace{-1mm}, \IEEEmembership{Member, IEEE}
\thanks{This work is partially supported by the National Natural Science Foundation of China (No: 61773197), the Shenzhen Fundamental Research Program (No: JCYJ20200109141622964), 
the Intel ICRI-IACV Research Fund ($CG\#52514373$). 
(Corresponding authors: Gongjin Lan; Qi Hao.)}
\thanks{Gongjin Lan, Yu Wu, Qi Hao are with the Department of Computer Science and Engineering, Southern University of Science and Technology, Shenzhen, 518055, China (e-mail: langj@sustech.edu.cn, wuy@mail.sustech.edu.cn, hao.q@sustech.edu.cn)}
\thanks{Fei Hu is with the Department of Electrical and Computer Engineering, University of Alabama, Tuscaloosa, AL (email: fei@eng.ua.edu)}
}



\markboth{IEEE TRANSACTIONS ON HUMAN-MACHINE SYSTEMS, VOL. 53, NO. 1, FEBRUARY 2023}%
{Shell \MakeLowercase{\textit{et al.}}: Bare Demo of IEEEtran.cls for IEEE Journals}

\maketitle

\begin{abstract}

Human Pose Estimation (HPE) has attracted a significant amount of attention from the computer vision community in the past decades.
Moreover, HPE has been applied to various domains such as human-computer interaction, sports analysis, and human tracking via images and videos.
Recently, deep learning-based approaches have shown state-of-the-art performance in HPE-based applications.
Although deep learning-based approaches have achieved remarkable performance in HPE, 
a comprehensive review of deep learning-based HPE methods remains lacking in the literature.
In this paper, we provide an up-to-date and in-depth overview of the deep learning approaches in vision-based HPE. 
We summarize these methods of 2D and 3D HPE, and their applications, discuss the challenges and the research trends through bibliometrics and provide insightful recommendations for future research. 
This article provides a meaningful overview as introductory material for beginners to deep learning-based HPE, as well as supplementary material for advanced researchers.
\end{abstract}

\begin{IEEEkeywords}
Human pose estimation, Human performance assessment, Deep learning, Action recognition, Bibliometric.
\end{IEEEkeywords}

%
\IEEEpeerreviewmaketitle

\section{Introduction}
\label{sec:introduction}

Human Pose Estimation (HPE) refers to estimating the positions of human joints and their associations in images or videos, which is a popular research topic in computer vision.
It has been widely applied to various applications, such as action analysis \cite{liu2020disentangling},
HCI \cite{shotton2011real}, gaming \cite{ke2010real}, sport analysis \cite{wang2019ai,park2017accurate}, motion capture \cite{yang2020transmomo}, computer-generated imagery \cite{hornung2007character,willett2020pose2pose}. 
Although HPE has been studied for decades, it is still an open and challenging task since the two main aspects of the wide diversity of the human body (such as various human poses, various clothing, environment or illumination conditions) and reconstruction ambiguity caused by occlusions (particularly the crowd) \cite{toshev2014deeppose,sun2019deep}.

The early HPE approaches often use predefined models and statistical learning to describe the human poses \cite{hogg1983model,moeslund2001survey}. 
However, those methods are incapable of learning from a large amount of data and suffer from limited model representation capability. 
In recent years, deep learning-based approaches yield great improvements in many computer vision tasks such as classification \cite{krizhevsky2012Imagenet}, object detection \cite{lan2018real,lan2019evolving}, and HPE \cite{toshev2014deeppose}. 
The success of deep learning in HPE is mainly due to the following three facts: the availability of big data, superior representation capability of deep neural networks, and high-performance hardware (e.g., GPU platform).
The deep learning-based methods dramatically outperform the traditional approaches.

Although there are many promising methods in the vision-based HPE via deep learning models 
\cite{toshev2014deeppose,andriluka2018posetrack,mcnally2021evopose2d}, 
a lack of articles with an up-to-date and in-depth review of this domain.
We emphasize that a comprehensive overview of HPE should cover both 2D and 3D HPE studies.
In this paper, we aim to provide a complete and solid survey, analyze the research challenges, and point out the research trends in HPE. 
In particular, we apply bibliometrics to retrieve scientific publications for analyzing the research trends in HPE. 
This paper comprehensively reviews HPE topics that cover both 2D and 3D HPE studies, discusses the challenges, observes the trends, and provides detailed bibliometrics.

\subsection{Related Work}
\label{subsec:related}
To date, several survey papers have discussed the related studies in HPE. 
Dang et al. \cite{dang2019deep} provided a survey on deep learning-based 2D HPE, including single- and multi-person pipelines.
Poppe et al. \cite{poppe2010survey} presented an overview of the literature on vision-based human action recognition. 
Currently, there are many studies that investigate monocular HPE, and several survey papers that discuss the studies in monocular HPE. 
In \cite{chen2020monocular,liu2022recent,gong2016human}, the studies of monocular HPE are reviewed comprehensively, particularly the deep learning-based methods.
Dang et al. \cite{dang2020sensor} provided a comprehensive survey on sensor- or vision-based human activity recognition.
Gadhiya et al. \cite{gadhiya2021analysis} analyzed and compared several prevalent HPE methods.
The latest work \cite{wang2021deep} reviewed the studies of deep learning-based 3D HPE.
Although these survey papers covered HPE-related topics, they focus on one of the specific topics in HPE.
In this paper, we aim to provide a comprehensive survey on the vision-based HPE based on deep learning models in terms of 2D and 3D HPE.
\subsection{Contributions}
\label{subsec:contributions}

In this paper, we aim to provide a detailed overview of the existing studies on deep learning-based human pose estimation.
This review has three objectives:
\begin{itemize}
    \item Delineate the picture of this field from a ‘helicopter view’.
    \item Clarify main research streams and provide a complete overview of vision- and deep learning-based HPE.
    \item 
    Discuss the challenges and the research trends through bibliometrics, and provide insightful recommendations for future research. 
\end{itemize}

\begin{figure}[!ht]
    \centering \small
    \includegraphics[width=\textwidth]{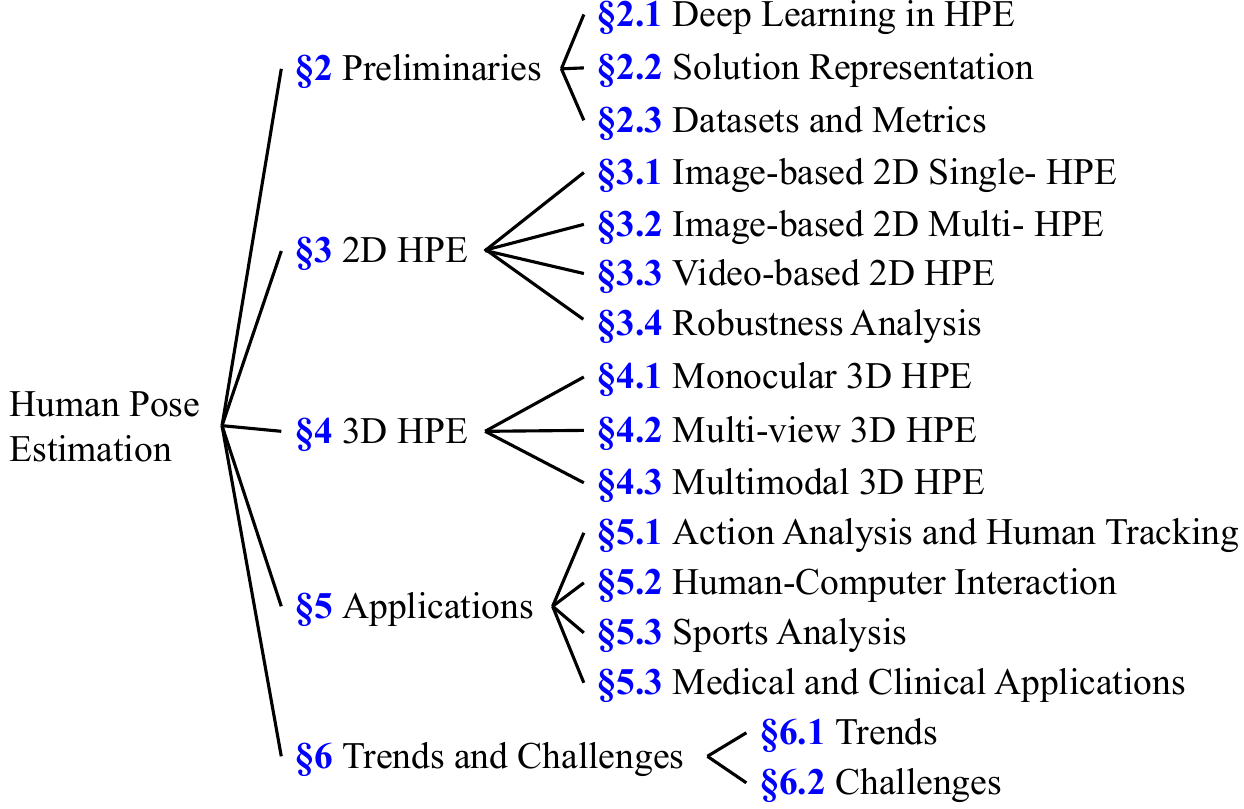}
    \caption{Taxonomy of this survey.}
    \label{fig:taxonomy}
\end{figure}

This survey covers both 2D- and 3D-based approaches. 
\autoref{fig:taxonomy} shows the taxonomy of the approaches (e.g., image-based or video-based, 2D HPE or 3D HPE, monocular or multi-view), applications, trends, and challenges in this survey. These contributions provide our survey with a more solid, up-to-date, and in-depth insight than the existing survey papers.

The rest of this paper is organized as follows.
In \autoref{sec:def}, we present the preliminary knowledge, common datasets, and the metrics for HPE.
Image-based 2D HPE is summarized in \autoref{sec:image-based2DHPE}. 
We address video-based 2D HPE in \autoref{sec:videobase}. 
The studies of 3D HPE is addressed in \autoref{sec:3dHPE}. 
The applications of vision-based HPE using deep learning are presented in \autoref{sec:applications}.
Finally, the research trends and challenges are discussed in \autoref{sec:trends_chall}, followed by the conclusions in \autoref{sec:conclusion}.

\section{Preliminary Knowledge}
\label{sec:def}

In this section, we introduce the preliminary knowledge, including essential concepts to guide the readers on the big picture of HPE and describe the solution representation and the well-known datasets with the performance metrics.

\begin{table*}[!ht]
\centering \footnotesize
\renewcommand{\arraystretch}{1.0}
\setlength\tabcolsep{3pt}
\begin{tabular}{llll 
l lcl}
\toprule
2D/3D & Type & Year & Dataset  & URL (Open dataset) & Data Scale &\#Joints & Metrics  \\ \midrule
\multirow{8}{*}{2D} & \multirow{4}{*}{\makecell{Single-\\Person}} &  2010 & LSP \cite{Johnson10LSP} & \url{http://sam.johnson.io/research/lsp.html} & 2K images & 14& PCK\&PCP \\
& &  2013 & FLIC \cite{modec13FLIC} & \url{https://bensapp.github.io/flic-dataset.html} & 5K images & 10 & PCK\&PCP \\
& &  2013 & J-HMDB \cite{Jhuang:ICCV:2013JHMDB} & \url{http://jhmdb.is.tue.mpg.de/} & 928 video clips &15 & PCK\&PCP \\
& &  2013 & PennAction \cite{zhang2013actemesPenn} & \url{http://dreamdragon.github.io/PennAction/} & 2326 video clips & 13 & PCK\&PCP \\ \cline{2-8}
& \multirow{4}{*}{\makecell{Multi-\\Person}} 
& 2014 & MPII \cite{andriluka14cvprmpii} & \url{http://human-pose.mpi-inf.mpg.de/} & 25K images \& 40K persons & 16 & mAP \\
& & 2016 & COCO \cite{lin2014microsoft} & \url{https://cocodataset.org/\#home} & 330K images \& 250K persons & 17 & AP \& AR \\
& & 2018 & PoseTrack \cite{andriluka2018posetrack} & \url{https://github.com/umariqb/PoseTrack-CVPR2017} & 46K frames \& 276K persons & 15 & mAP \\
& & 2019 & CrowdPose \cite{li2019crowdpose} & \url{https://github.com/Jeff-sjtu/CrowdPose} & 20K images \& 80K persons & 14 & mAP \\
\midrule
\multirow{3}{*}{3D} & \makecell[l]{Single-\\Person} & 2014 & Human3.6M \cite{h36m_pami} & \url{http://vision.imar.ro/human3.6m/} & 3.6M frames \& 4 camera views & 17 & MPJPE \\
\cline{2-8}
& \multirow{4}{*}{\makecell[l]{Multi-\\Person}} & 2017 & CMU Panoptic \cite{joo2017panoptic} & \url{http://domedb.perception.cs.cmu.edu/} & 1.5M frames \& 512 camera views & 15& MPJPE \\
& & 2018 & 3DPW \cite{von2018recovering} & \url{https://virtualhumans.mpi-inf.mpg.de/3DPW/} & 51K frames \& 1 camera view & 18&MPJPE \\
& & 2017 & MPI-INF-3DHP \cite{mehta2017monocular} & \url{https://vcai.mpi-inf.mpg.de/3dhp-dataset/} & 1.3M frames \& 16 camera views &15 & MPJPE \\
& & 2018 & JTA \cite{fabbri2018learning} & \url{https://github.com/fabbrimatteo/JTA-Dataset} & 460K frames \& 1 camera view & 14& MPJPE\\
\bottomrule
\end{tabular}
\caption{Illustration of the well-known datasets in 2D and 3D HPE.}
\label{tab:2d_dataset}
\end{table*}

\subsection{Deep Learning in HPE}
Currently, deep learning-based approaches have become state-of-the-art methods in HPE. 
The availability of big datasets, advanced hardware like GPU, and the surpassing performance of deep neural networks lead to the increasing interest in deep learning-based HPE. 
In this subsection, we discuss three topical types of neural networks in HPE: Convolutional Neural Networks (CNNs), Recurrent Neural Networks (RNNs), and Graph Convolutional Networks (GCNs). 

\subsubsection{Convolutional Neural Networks} \hfill

In general, a CNN for HPE tasks consists of two parts. 
The first part commonly uses off-the-shelf generic pre-trained networks such as ResNet \cite{he2016deepresidual} to extract features, the so-called backbone network. 
The second part, called the prediction head, predicts human poses with the extracted features. 

The well-known networks like AlexNet \cite{krizhevsky2012Imagenet} and ResNet \cite{he2016deepresidual} show remarkable classification performance on the dataset of ImageNet, and advanced performance in HPE \cite{fang2017rmpe,cao2017realtime} as well.
However, there is a gap between classification and HPE tasks since their targeted features and prediction differ from each other.
Instead of directly using the backbones from classification tasks, HPE-specified backbones need to be improved for HPE tasks. 
For example, Hourglass \cite{newell2016stacked}, Cascaded Pyramid Network (CPN) \cite{chen2018cascaded}, and HRNet \cite{sun2019deep} are proposed to be the backbone for the deep learning-based approaches in HPE.

For prediction heads, there are mainly two representative solutions in HPE. 
The one directly predicts joint coordinates, which is regarded as the regression paradigm. 
The other one generates an intermediate heatmap representation before computing joint coordinates.
For the regression paradigm, fully connected layers are often adopted to regress concrete keypoint coordinates. 
For the heatmap prediction paradigm, the operation of upsampling \cite{chen2018cascaded,newell2016stacked,xiao2018simple} is generally used to generate higher resolution heatmaps.

\subsubsection{Recurrent Neural Networks} \hfill

Recurrent neural networks stake temporal information among sequential inputs into consideration. 
They are widely used in video-based HPE by considering videos as sequential RGB images. 
We present the general pipeline of this type of networks in \autoref{fig:video_based_2D_HPE_pipeline}. 
RNN-based methods perform robust and the state-of-the-art accuracy \cite{luo2018lstm,artacho2020unipose}. 

\subsubsection{Graph Convolutional Networks} \hfill

Instead of taking images as input, graph convolutional networks (GCNs) take graphs as input.
As a human skeleton can be naturally represented as a graph, GCNs-based methods are prevalent in many skeleton-based tasks.
In HPE, GCNs are generally expected to better exploit the relationship among keypoints and used for the pose refinement \cite{bin2020structure}, joint association \cite{qiu2020dgcn}, 2D-to-3D pose lifting \cite{zhao2019semantic,hu2021conditional,cai2019exploiting}.


\subsection{Pose Representation}
Although the natural representation of human poses (keypoint positions) uses coordinates in the form of ordered tuples, the existing studies have shown a significant improvement by representing solutions with heatmaps that can be regarded as a confidence map. 
Currently, the heatmap representation has become a prevalent solution representation in HPE. 
In this subsection, we describe how heatmap representation works in the single-person and multi-person HPE.

\subsubsection{Heatmaps in Single-person HPE} \hfill

To enable neural networks for the heatmap prediction of human joints, the ground truth of a heatmap is essential.
An intuitive way of producing the ground truth is to design a probability heatmap for each keypoint. 
For example, for single-person pose estimation tasks, the heatmaps of {\small $\mathcal{K}$} keypoints can be defined as {\small $\mathcal{K}$} matrices with the identical size to the input image $x$. 
The value of position $p$ (noted as {\small $\mathcal{H}_k(p)$}) can be generated by a 2D Gaussian distribution centered at the position of joint $k$ in $x$.
\begin{equation} \small
    \label{eq: heatmap}
    \mathcal{H}_k(p) = e^{\frac{\|p-p_{k}^*\|_{2}^2}{\sigma^2}}, \forall k=1,2,\ldots,\mathcal{K}
\end{equation}
where $p_k^*$ is the position of joint $k$ in $x$.
In a heatmap, the predicted position (noted as $p_{pred}^*$) can be calculated by regression models as:
\begin{equation} \small
    p_{pred}^* =\sum_{p \in \mathcal{P}} \mathcal{H}_k(p)*p,
\end{equation}
where {\small $\mathcal{P}$} is the set of the possible positions of joint $k$.

\subsubsection{Heatmaps in Multi-person HPE} \hfill

There are two primary patterns of heatmaps in multi-person scenarios. 
One is to generate {\small $\mathcal{N}\times \mathcal{K}$} heatmaps for {\small $\mathcal{N}$} persons and their {\small $\mathcal{K}$} joints separately. 
The other is to generate {\small $\mathcal{K}$} heatmaps for {\small $\mathcal{K}$} joint of all persons.
For the former pattern ({\small $\mathcal{N}\times \mathcal{K}$} heatmaps), each value at the position $p$ in a heatmap (noted as $\mathcal{H}_{k,n}(p)$) can be calculated as: 
\noindent
\begin{equation} \small
    \label{eq:multi-personHeatmap}
    \mathcal{H}_{k,n}(p) = e^{\frac{\|p-p_{k,n}^*\|_{2}^2}{\sigma^2}}, ~ \forall k=1,2,\ldots,\mathcal{K}, ~ \forall n=1,2,\ldots,\mathcal{N}
\end{equation}
where {\small $\mathcal{K}$} and {\small $\mathcal{N}$} are the numbers of joints and persons, respectively.
For the latter pattern, a general method to produce the ground truth heatmaps is aggregating {\small $\mathcal{N}$} single-person heatmaps into one heatmap by using a $\max$ operator:
\begin{equation} \small
    \label{eq: multi-personAggo}
    \mathcal{H}_k(p) = \max_{n} \mathcal{H}_{k,n}(p), \forall n=1,2,\ldots,\mathcal{N}
\end{equation}
In summary, \autoref{eq:multi-personHeatmap} and \autoref{eq: multi-personAggo} are the two main methods to calculate the heatmaps in multi-person HPE.

\subsection{Datasets and Metrics}
Datasets are critical for training and evaluating neural networks for deep learning-based HPE. 
In this subsection, we introduce the popular datasets and their applicable tasks, then review the evaluation metrics for image-based 2D HPE, video-based 2D HPE, and 3D HPE.

\subsubsection{Datasets in 2D HPE}
\label{subsubsec:datasets} \hfill

Many datasets have been proposed to evaluate the performance of 2D HPE approaches. 
Here, we introduce the prevalent datasets used in 2D HPE. 
Early datasets like LSP \cite{Johnson10LSP}, FLIC \cite{modec13FLIC}, Penn Action \cite{zhang2013actemesPenn}, and J-HMDB \cite{Jhuang:ICCV:2013JHMDB} mainly focus on single-person scenes with relatively small scales.
The recent datasets such as COCO \cite{lin2014microsoft}, MPII \cite{andriluka14cvprmpii}, CrowdPose \cite{li2019crowdpose}, and PoseTrack \cite{andriluka2018posetrack} are used for multi-person HPE with larger-scale data. 
We summarize and provide the links of the prevalent datasets as well as their scales and the evaluation metrics in \autoref{tab:2d_dataset}.

\subsubsection{Datasets in 3D HPE} \hfill

Unlike the datasets in 2D HPE, acquiring accurate 3D annotations for human joints in 3D HPE often requires a motion capture system that is generally hard to be installed in the outdoor environment.
Most 3D HPE datasets are created in the indoor environments or simulation, such as CMU Panoptic \cite{joo2017panoptic}, 3DPW \cite{von2018recovering}, MPI-INF-3DHP \cite{mehta2017monocular}, JTA \cite{fabbri2018learning}.
Here we introduce the well-known datasets in 3D HPE and summarize their characteristics in \autoref{tab:2d_dataset}.

\subsubsection{Metrics in 2D HPE} 
\label{subsub:metrics_2D}
\hfill

In scientific research, we usually need metrics to evaluate how well a method performs. 
Here we introduce two common metrics used in HPE.
\emph{Percentage of Correct Keypoints (PCK)}
literally indicates the percentage of correct detected keypoints, which can be noted as {\small $\mathcal{PCK}=k/\mathcal{N}$}, where k is the number of correct predicted keypoints, {\small $\mathcal{N}$} is the total number of keypoints.
This metric is generally applied to the studies with the LSP dataset, MPII dataset, and FLIC dataset in the early 2D HPE methods (see \autoref{tab:SPPE}).

\begin{table*}[!ht]
\centering \footnotesize
\renewcommand{\arraystretch}{1.0}
\setlength\tabcolsep{4pt} 
\begin{tabular} {
l l l c l c l l}
\toprule
Studies & Years & Backbone & Input Size & Highlights&  PCKh@0.5 & \#Params & GFlops \\ \midrule
Toshev \& Szegedy \cite{toshev2014deeppose} & 2014 & AlexNet & $256 \times 256$  &  Original deep learning-based method for HPE & -& - & - \\
Tompson et al. \cite{tompson2014joint} & 2014 & AlexNet & $320\times 240$  &  Utilization of heatmaps for solution representation & 82.0& - & - \\
Wei et al. \cite{wei2016convolutional} & 2016 & CPM & $368\times 368$  &  A convolutional pose machine & 88.5& 31.23M & 85.0 \\
Newell et al. \cite{newell2016stacked} & 2016 & Hourglass & $256\times 256$  &  Stacked Hourglass Modules & 90.9& 23.7M & 41.2 \\
Xiao et al. \cite{xiao2018simple} & 2018 & ResNet & $256\times 256$  &  A simple yet effective architecture for HPE& 90.2& 68.64 M & 17.02 \\
Chu et al. \cite{chu2017multi} & 2017 & Hourglass & $256\times 256$  &  Attention mechanism in contextual representations & 91.5 & 58.1M & - \\
Yang et al. \cite{yang2017learningFeaturePyramid} & 2017 & Hourglass & $256\times 256$ & Pyramid Residual Module & 92.0& 26.9M & 45.9 \\
Sun et al. \cite{sun2019deep} & 2019 & HRNet & $256\times 256$  &  HRnet for high-resolution representations & 92.3& 28.54M & 10.27 \\
Bulat et al. \cite{bulat2020toward}  & 2020 &Hourglass+UNet & $256\times 256$  &  A hybrid structure by combining \cite{newell2016stacked} and U-Net \cite{rafi2016efficient} & 94.1& 8.5M & 9.9 \\
\bottomrule
\end{tabular}
\caption{The state-of-the-art approaches for 2D single-person pose estimation. The PCKh@0.5 scores are shown in the last column and defined in \autoref{subsub:metrics_2D}. They  are obtained by testing the methods on the MPII dataset.}
\label{tab:SPPE}
\end{table*}

\emph{Object Keypoint Similarity (OKS)}
was proposed originally in the COCO competition \cite{lin2014microsoft} as a variable to compute mean Average Precision (mAP).
It can be calculated as follows:
\begin{equation} \small
    \mathcal{OKS} = \frac{\sum_{i}(-d_i^2/2s^2k_i^2)\delta_{(v_i>0)}}{\sum_{i}\delta_{(v_i>0)}}, \quad s.t. ~ v_i \in \{0,1,2\}
\end{equation}
where $i$ is a joint index, $d_i$ is the distance between the predicted joint and ground truth, $s$ (object scale) and $k_i$ are the keypoint constants given by the COCO dataset, $\delta = 1$ when $v_i > 0$, otherwise $\delta = 1$, and $v_i$ is the visibility flag ($v_i = 0$: not labelled, $v_i = 1$: labelled but not visible, and $v_i = 2$: labelled and visible) of the ground truth. 
The average precision (AP) can be calculated with the {\small $\mathcal{OKS}$} value by: 
\begin{equation} \small
    AP = \frac{\mathcal{TP}_{(\mathcal{OKS}>td)}}{\mathcal{TP}_{(\mathcal{OKS}>td)}+\mathcal{FP}_{(\mathcal{OKS}\leq td)}}
\end{equation}
where {\small $\mathcal{TP}$} and {\small $\mathcal{FP}$} are the numbers of true positive and false positive respectively, $td$ is a {\small $\mathcal{OKS}$} threshold.
The mean AP (mAP) is the mean of AP over ten {\small $\mathcal{OKS}$} thresholds at (0.50, 0.55, . . . , 0.90, 0.95) \cite{lin2014microsoft}, which is a common metric to evaluate 2D multi-person pose estimation (see \autoref{tab:MPPE}).

\subsubsection{Metrics in 3D HPE} \hfill
\label{para:mpjpe}

Generally, the metrics of PCK can be extended to evaluate 3D HPE. 
However, \emph{Mean Per Joint Position Error (MPJPE)} is currently a popular metric in 3D HPE (see \autoref{tab:image-based3dhpe}).
$\text{MPJPE}$ calculates the euclidean distance between the predicted joint coordinates and ground truth joint coordinates, which can be formulated as:
\begin{equation} \small
\label{eq:mpjpe}
    \text{MPJPE} = \frac{1}{\mathcal{T}\cdot \mathcal{N}}\sum\limits_{t=1}^\mathcal{T}\sum\limits_{i=1}^{\mathcal{N}} \left\lVert J_i^{t}-J_{root}^t - (\hat{J_i}^t-\hat{J}_{root}^t)\right\rVert^2
\end{equation}
Where $i$ and $t$ are the indexes of a joint and a sample respectively; $J_i$ and $\hat{J_i}$ refer to the predicted coordinate and ground-truth coordinate of $i-th$ joint. 
$J_{root}$ refers to the coordinate of the root joint, which is usually predefined as the human pelvis.

\section{2D Human Pose Estimation}
\label{sec:image-based2DHPE}
The image-based 2D HPE is to estimate human joint positions in images. 
Early approaches mainly use model-based methods. 
Currently, deep learning-based methods have shown superior performance in 2D HPE \cite{toshev2014deeppose}.
In this section, we introduce the deep learning-based 2D HPE approaches from three aspects: image-based single-person pose estimation (SPPE), image-based multi-person pose estimation (MPPE), and video-based 2D HPE.
Finally, we summarize the open-source codes of the state-of-the-art 2D HPE in \autoref{tab:open-source}.

\subsection{Image-based 2D Single-Person HPE}
\label{sec:imageSingleperson}

The SPPE task is to estimate the pose of a single person in an image. 
It is a fundamental task in HPE and is often used as a basic component of other HPE tasks. 
For example, the well-known study \cite{toshev2014deeppose} proposed a cascaded multi-stage neural network to predict and refine human poses, in which images are cropped to ensure a single person in each image. 
To our knowledge, it is the original work that applies deep neural networks (DNN) for the HPE.

Recent DNN-based approaches dramatically improve the accuracy of SPPE. 
Many generic neural networks with efficient architecture show remarkable performance in various applications \cite{krizhevsky2012Imagenet,he2016deepresidual}.
There are mainly two ways to improve SPPE performance: improving the solution representation and designing advanced neural networks. 
In \cite{tompson2014joint,jain2014modeep,zhang2020distribution}, the methods with solution representation significantly improved the SPPE performance. 
In addition, well-designed neural architectures \cite{sun2019deep,wei2016convolutional} also performed remarkable performance in HPE. 
We summarize the state-of-the-art approaches for 2D single-person pose estimation in \autoref{tab:SPPE}.

\subsubsection{Solution Representation} \hfill

In general, there are mainly two types of solution representations. 
One is to represent joint positions by using coordinates. 
The other one is to represent joint positions in the form of probability distributions, i.e., heatmaps as represented in \autoref{eq: heatmap}. 
Early studies mainly used a DNN-based regression from the input images to the estimated coordinates directly. 
DeepPose \cite{toshev2014deeppose} is an early classic study that applied DNNs to represent coordinate regression from RGB images and multi-stage refinement for the SPPE tasks.

Solution representation of heatmaps significantly improves the performance of DNN-based methods in HPE \cite{rafi2016efficient}. 
The methods of heatmap generation have been widely used in HPE tasks since heatmaps were proposed by Tompson et al. \cite{tompson2014joint} and Jain et al. \cite{jain2014modeep}.
Moreover, Zhang et al. \cite{zhang2020distribution} improved the heatmap representation by the distribution-aware coordinate representation of keypoints (DARK). 

\subsubsection{Neural Network Design} \hfill

How to design the architecture of neural networks in HPE is a crucial topic.
A generic way is to use the existing well-known neural networks (e.g., AlexNet \cite{krizhevsky2012Imagenet}, ResNet \cite{he2016deepresidual})
that have already shown superior performance in other computer vision tasks. 
For example, Toshev et al. \cite{toshev2014deeppose} used an AlexNet-like to regress the coordinates of human joints from images. 
Xiao et al. \cite{xiao2018simple} used ResNet as the backbone of neural networks for HPE.

The latest neural network, the transformer that adopts the attention mechanism, has been applied in HPE with superior performance. 
In general, transformers process the features extracted by CNNs and leverage the attention mechanism to automatically capture long-range relationships among features.
In 2021, Mao et al.\cite{mao2021tfpose} used ResNet to extract features from images and a Transformer model to predict keypoint positions.
Li et al.\cite{li2021poseTransformer} utilized vision transformers to implement regression-based HPE, where two cascaded transformers are applied to predict the bounding boxes of the person as well as their keypoints positions. 
 Yang et al. \cite{yang2021transpose} proposed a new network model, TransPose which introduces the transformer for human pose estimation. 
The attention layers built-in transformers enable TransPose to capture long-range relationships efficiently.
In addition, Xu et al. proposed a scalable HPE backbone, VitPose \cite{xu2022vitpose} that employs plain and non-hierarchical vision transformers as backbones to extract features for a given person instance and a lightweight decoder for pose estimation.
To date, the existing survey papers have not discussed the studies with transformers in HPE.

Another generic way is to design specific networks for HPE tasks.
For example, Newell et al. \cite{newell2016stacked} designed a stacked hourglass network as the backbone for exacting multi-scale features automatically. 
Hourglass Networks are used in many HPE studies as backbone networks. 
Newell et al. \cite{newell2017associative} used the Hourglass Network backbone to estimate multi-person poses. 
BlazePose \cite{bazarevsky2020blazepose} utilized an Hourglass-based encoder network to accomplish real-time single-person pose estimation driven by the MediaPipe framework \cite{lugaresi2019mediapipe}, which achieved a speed of 30 frames per second on the Google Pixel 2 mobile phone.
Bulat et al. \cite{bulat2020toward} designed a hybrid network structure with a combination of the hourglass network and U-Net \cite{rafi2016efficient}. 
Chen et al. \cite{chen2018cascaded} presented another remarkable study that a cascaded pyramid network is proposed to concatenate features among different scales. 
In addition, many studies enhanced the model performance by slightly modifying the units of a backbone. 
Specifically, Chu et al. \cite{chu2017multi} incorporated the attention mechanism into the hourglass network. 
Yang et al. \cite{yang2017learningFeaturePyramid} proposed a pyramid residual module to replace the residual module in the hourglass network.


\subsection{Image-based 2D Multi-Person HPE}
\label{sec:imageMultiperson}

In general, MPPE is a more challenging task than SPPE because of the high complexity in terms of solution space and mutual occlusions.
\begin{figure}[!htbp]
    \centering
    \large
    \begin{adjustbox}{max width=0.95\columnwidth}
    \begin{tabular}{c c c}
        \includegraphics[width=0.49\textwidth]{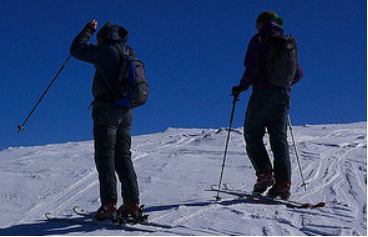}& \hspace{-0.6cm}
        \includegraphics[width=0.49\textwidth]{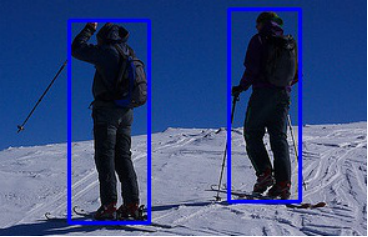}& \hspace{-0.6cm}
        \includegraphics[width=0.49\textwidth]{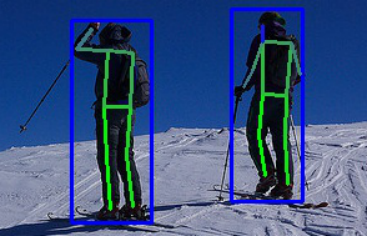} \\
        Input Image    &  \hspace{-0.5cm} Detected Bounding Boxes &  Human Poses  \\
        \includegraphics[width=0.49\textwidth]{images/0.png}& \hspace{-0.6cm}
        \includegraphics[width=0.49\textwidth]{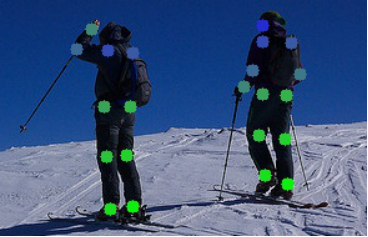}& \hspace{-0.6cm}
        \includegraphics[width=0.49\textwidth]{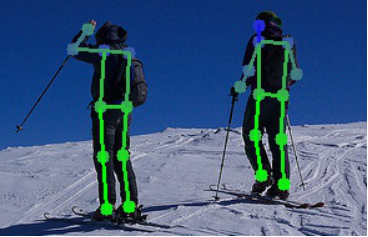} \\
         Input Image    & \hspace{-0.5cm} Detected Keypoints  &    Human Poses    \\
    \end{tabular}
    \end{adjustbox}
    \caption{Illustration of the top-down and bottom-up framework. The first row shows a representative top-down framework with two stages: proposing a bounding box for each person and estimating human poses. The second row shows a representative bottom-up framework with two stages: detecting the joints of persons and grouping the joints into an associated person. The original images are from COCO dataset \cite{lin2014microsoft}.}
    \label{fig:MPPE_pipeline}
\end{figure}
The current solutions of MPPE generally fall into either top-down approaches or bottom-up approaches:
\begin{itemize}
    \item \textbf{Top-down} approaches apply a person detector to detect all single-persons, followed by estimating the joints of each single-person and calculating each single-person pose separately.
    \item \textbf{Bottom-up} approaches detect all joints in an image, followed by associating/grouping the joints into an associated person.
\end{itemize}
In this subsection, we address the top-down and bottom-up approaches and discuss their merits and flaws.
We illustrate the frameworks of both approaches in \autoref{fig:MPPE_pipeline} and summarize the recent well-known MPPE approaches in \autoref{tab:MPPE}.

\begin{table*}[!ht]
\centering \footnotesize
\renewcommand{\arraystretch}{1.0}
\setlength\tabcolsep{2pt} 
\begin{tabular}{l l c l l m{6.3cm} c c c}
\toprule
Methods & Studies/Years & Input Size & Backbone & \makecell[l]{Detector/\\Grouping} & \multicolumn{1}{c}{Highlights} & mAP & \#Params & GFlops \\ \midrule
\multirow{11}{*}{Top-down} & \cite{he2017mask} / 2017 &  - & ResNet &  Mask R-CNN &  Detecting persons and keypoints by mask R-CNN & 63.1 & - & - \\
& \cite{fang2017rmpe} / 2017 & $320\times 256$ & PyraNet & Faster R-CNN & Addressing inaccurate bounding box of persons & 72.3 & 28.1M & 26.7 \\
& \cite{chen2018cascaded} / 2018  & $ 384\times 288 $ & CPN & FPN &  A two-stage cascaded network to refine keypoints & 72.1& 27.0M & 6.2 \\
& \cite{xiao2018simple} / 2018  & $ 384\times 288 $ & ResNet & Faster R-CNN &  A remarkable simple baseline & 72.2 & - & - \\
& \cite{sun2019deep} / 2019 & $384\times 288$ & HRNet & Faster R-CNN &  HRnet for high-resolution representations & 75.5 & 63.6M & 32.9 \\
& \cite{li2019rethinking} / 2019 &$384\times 288$ & MSPN & MegDet & A multi-stage network & 76.1 & 120M & 19.9 \\
& \cite{mcnally2021evopose2d} / 2020 & $512\times 384$ & EvoPose2D & Faster R-CNN & Neural Architecture Search & 76.8 & 14.7M & 17.7 \\
& \cite{bin2020structure} / 2020 & $384\times 384$ & ResNet & FPN & GCN-based keypoint refinement & 72.9 & 25.2M & 12.9\\
& \cite{mao2021tfpose} / 2021 & $384\times 288$ & \makecell[l]{ResNet+\\Transformer} & Faster R-CNN & Direct Coordinate Regression via Transformer & 72.2 & - & 20.4G \\
& \cite{yang2021transpose} / 2021 & $256\times 192$ & TransPose & Faster R-CNN & Transformer decoders with CNNs-based extractor & 75.8 & 17.5M & 21.8G \\
& \cite{xu2022vitpose} / 2022 & $256\times 192$ & VitPose & Faster R-CNN & Vision Transformer Backbone Baseline & 79.8 & 632M & - \\
\midrule
\multirow{8}{*}{Bottom-up} &  \cite{insafutdinov2016deepercut} / 2016 & $256\times 256$  & ResNet & \makecell[l]{Integer \\ Programming} & \makecell[l]{ResNet-based detectors and \\ image-conditioned pairwise terms} & - & - & - \\
&  \cite{cao2017realtime} / 2017 &$256\times 256$  & VGG+CPM & \makecell[l]{Part Affinity \\ Fields} & Grouping by body association and Hungarian algo. & 61.8 & 25.94M & 160 \\
&  \cite{newell2017associative} / 2017 & $512\times 512$ & Hourglass & \makecell[l]{Associative \\ embedding} & pixel-wise joint embedding for grouping & 66.3 & 277.8M  & 206.9 \\
&  \cite{nie2019single} / 2019 & $384\times 384$ & Hourglass & $\emptyset$ & Predicting root and joints position & 66.9 & - & - \\
& \cite{qiu2020dgcn} / 2020 & $641\times 641$ & ResNet & DGCN & Graph convolutional network for grouping & 68.8 & 234M & - \\
& \cite{jin2020differentiable} / 2020 & $ 512\times 512$
& Hourglass & \makecell[l]{Associative \\ embedding} & Graph Clustering for grouping & 68.3 & - & - \\
&  \cite{Kocabas_2018_ECCV} / 2018 & $480\times 480$ & ResNet+FPN & \makecell[l]{Person \\ Detection} & Pose Residual Network assigns keypoints to instances & 69.6 & - & - \\
& \cite{cheng2020higherhrnet} / 2020 & $640\times 640$  & HigherHRNet & \makecell[l]{Associative \\ embedding} & HigherHRNet for the scale variation challenge & 70.5 & 63.8M & 154.3  \\
\bottomrule
\end{tabular}
\caption{The state-of-the-art studies for image-based 2D multi-person HPE. The mAP scores are defined in \autoref{subsub:metrics_2D}, are obtained by testing on the COCO dataset.}
\label{tab:MPPE}        
\end{table*}

\subsubsection{Top-down approaches}  \hfill

The top-down method is an effective, popular method in 2D multi-person pose estimation tasks. 
By using a pretrained human detector to crop images in top-down approaches, the multi-person tasks can be converted into single-person tasks. 
The performance of top-down methods can be enhanced with the improvements of both human detector and single-person pose estimator. 
He et al. \cite{he2017mask} showed that multi-person pose estimation can be implemented by extending HPE tasks to the detection tasks. 
Fang et al. \cite{fang2017rmpe} proposed a regional multi-person pose estimation framework to improve the performance of the human detector. 
Furthermore, the improvement of single-person pose estimators shows benefits for multi-person pose estimation \cite{zhang2020distribution,papandreou2017towards,chen2018cascaded}. 

In general, the top-down approaches show advanced performance on datasets and can be easily implemented by combining the existing detectors and SPPE models. 
However, the computing of top-down approaches is significantly increasing over the number of detected persons, which limits its real-time performance for multiple-person scenarios. 
Therefore, the top-down approaches are hardly applied to the real-time HPE tasks in complex scenarios, particularly the crowd scenes. 


\subsubsection{Bottom-up approaches} \hfill

Instead of performing keypoints detection in the proposed bounding boxes, the bottom-up methods often consist of two parts: keypoints detection and keypoints grouping. 
For keypoints detection, body keypoints of all persons in an image are detected directly by the heatmaps described in \autoref{eq:multi-personHeatmap}.
for keypoints grouping, the detected keypoints need to be grouped into the single-person.
The deep neural networks are generally applied to assign keypoints to the proposed bounding boxes of the detected person. 
Newell et al. \cite{newell2017associative} introduced associative embedding to train neural networks for assigning keypoints to each person. 
Jin et al. \cite{jin2020differentiable} applied graph neural networks to group the detected joints.

Moreover, it is shown in \cite{cheng2020higherhrnet} that the bottom-up approaches are more robust than the top-down approaches when applied to crowded scenes, which is crucial for practical applications. 
However, the top-down methods achieve better performance in terms of accuracy, while the computing time inevitably increases as the number of detected persons increases. 
By contrast, the bottom-up approaches take relatively constant computing time for multi-person HPE \cite{cao2017realtime,nie2019single}, which is much less sensitive to the number of targeted persons. 
In conclusion, the bottom-up approaches are conducive to real-time multi-person pose estimation on the low-performance hardware platform.

\subsection{Video-based 2D HPE}
\label{sec:videobase}



\begin{figure}[!ht]
    \centering
    \includegraphics[width=0.95\textwidth]{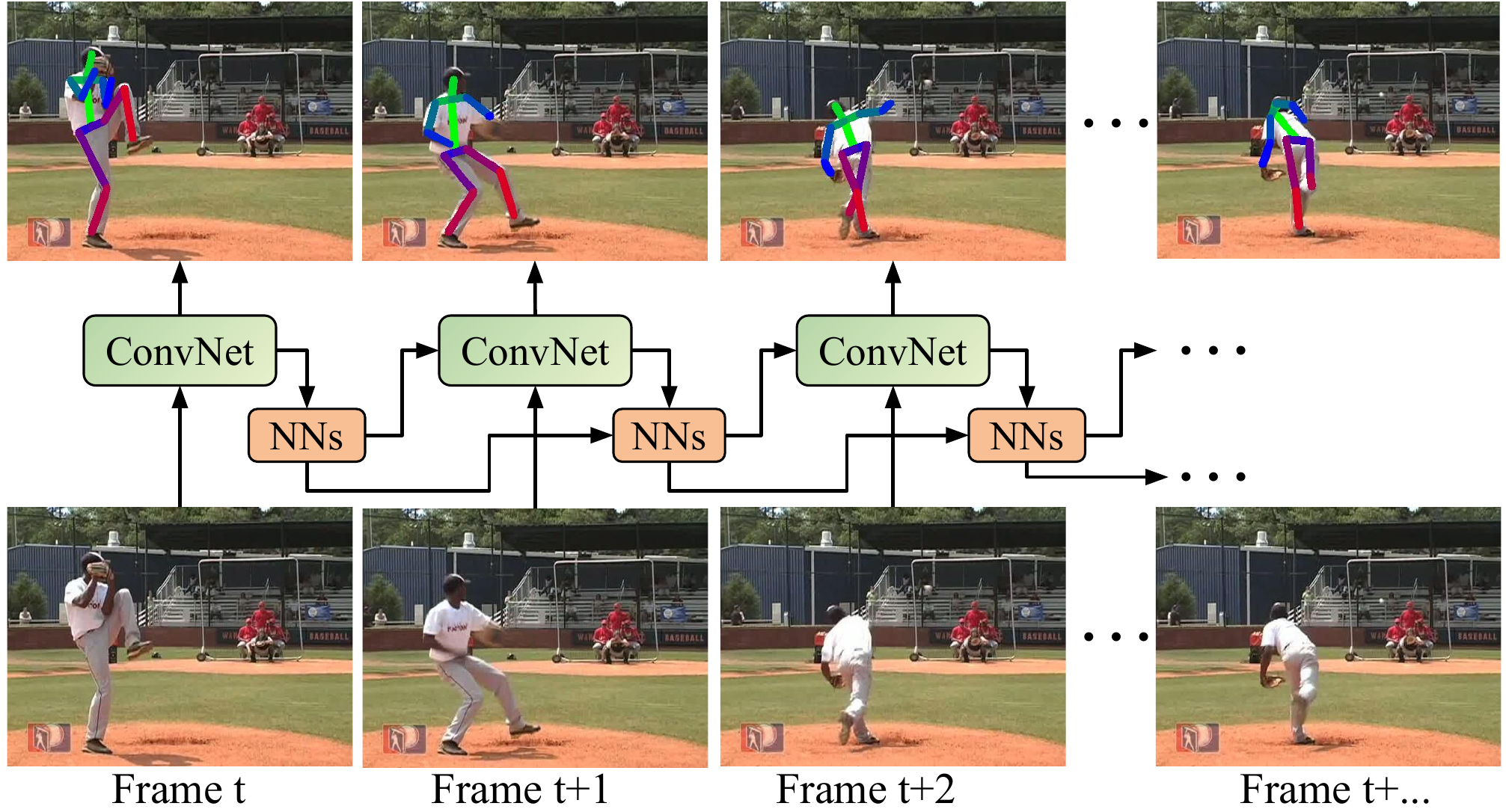}
    \caption{Illustration of a video-based 2D HPE pipeline where a Neural Networks-based (NNs-based) module is utilized to extract temporal information. 
    The original images are from the Penn Action dataset \cite{zhang2013actemesPenn}. }
    \label{fig:video_based_2D_HPE_pipeline}
\end{figure}

The video-based 2D HPE is generally a more complicated task than the image-based 2D HPE. 
A pipeline of the generic video-based 2D HPE is shown in \autoref{fig:video_based_2D_HPE_pipeline}. 
Unlike static images, video frames are likely to involve the problem of image degeneration such as motion blur and video defocus. 
Although the video-based 2D HPE performance may degenerate because of motion blur, video defocus, and temporary occlusions, these video-based approaches generally surpass image-based approaches by capturing temporal information. 
The correlations among video frames can be used to further improve the self-supervised approaches in HPE.

In video-based 2D HPE, it is costly to manually annotate human joints in each frame of video, which restricts the obtaining of large-scale datasets for the video-based HPE. 
To solve this issue, the temporal correlation and consistency among the sequences need to be fully investigated.
Jain et al. \cite{jain2014modeep} originally proposed a CNN-based approach to combine RGB images and motion features for improving both the accuracy and speed of HPE.
Specially, the recurrent neural networks, such as the well-known long short-term memory (LSTM), perform remarkably to capture temporal consistency among frames \cite{luo2018lstm,artacho2020unipose}. 
Nie et al. \cite{nie2019dynamic} proposed a dynamic kernel distillation model to distil previous temporal cues among frames. 
Although these methods perform state-of-the-art accuracy, they are supervised learning-based by using large-scale and densely labelled data in video-based HPE.
By contrast, semi-supervised learning is useful for video-based 2D HPE with sparsely labelled data since labelling large-scale and densely data is costly and labour-intensive. 
To this end, Bertasius et al. \cite{bertasius2019learning} proposed PoseWarper propagate pose information in sparsely labelled (each k frames) videos.

Finally, we collect the open-source implementations of well-known state-of-the-art 2D HPE works, as shown in \autoref{tab:open-source}. 
We summarize the prevalent real-time open-source implementations for practical applications (e.g., OpenPose, AlphaPose, a lightweight version of OpenPose, 
and BlazePose \cite{bazarevsky2020blazepose} driven by the framework of MediaPipe \cite{lugaresi2019mediapipe}
), which generally offer open-source codes to users.


\subsection{Robustness Analysis}

Robustness is a crucial property that should be considered in deep learning-based methods. 
We review the studies of robustness analysis for HPE in this subsection. 
Currently, robustness analysis is usually applied to 2D HPE but rarely considered in 3D HPE. 
Deep learning-based methods are generally sensitive and vulnerable to the attack of adversarial samples.

Although the robustness of HPE is rarely studied, it is crucial to be considered in the design and evaluation of HPE methods.
In \cite{newell2016stacked}, it is demonstrated that the PCK of the Hourglass network dramatically decrease from $89.4$ to $0.57$ in the testing with adversarial samples.
Wang et al. \cite{wang2021human} proposed the new datasets COCO-C, MPIIC, and OCHuman-C, which were reconstructed on the basis of COCO \cite{lin2014microsoft}, MPII \cite{andriluka14cvprmpii}, and OCHuman \cite{zhang2019pose2seg} for evaluating the robustness of HPE methods.
Shah et al. \cite{jain2019robustness} comprehensively investigated the adversarial robustness of HPE methods.
The experimental results show that 1) heatmap-based methods perform more robust than regression-based methods, and 2) different body joints generally perform different robustness to the attacks of adversarial samples.
For example, the head and neck exhibited prominent robustness, while the joints of the hips, knees, and ankles are sensitive to disturbances.
These works revealed the robustness of the existing deep learning-based HPE methods.

\section{3D Human Pose Estimation}
\label{sec:3dHPE}

The 3D HPE is to predict the 3D positions of human joints. 
It is a challenging task due to the large solution space and inherent ambiguity.
Moreover, the lack of outdoor large-scale 3D HPE datasets challenges the practical performance of 3D HPE approaches since the existing datasets 3D HPE is mostly collected by indoor motion capture systems
\cite{h36m_pami,joo2017panoptic}. 
Early studies either implemented 3D HPE with model-based approaches \cite{andriluka2012discriminative,bergtholdt2010study} or regarded the task as a regression problem which can be solved by optimization algorithms. 
Since the DNN-based methods \cite{li20143d,Ching20173D} outperform the previous works by automatically learning representations from large-scale data, deep learning-based methods have become the most popular methods in 3D HPE as well. 

In this section, we review deep learning-based 3D HPE studies as shown in \autoref{tab:image-based3dhpe}. 
According to the characteristics of inputs, we categorize the 3D HPE into three types: monocular 3D HPE, multi-view 3D HPE, and multimodal 3D HPE. 
Finally, we present a collection of the open-source code of the state-of-the-art 3D HPE approaches in \autoref{tab:open-source}. 




\begin{table*}[!ht]
\centering \small
\renewcommand{\arraystretch}{1.0}
\setlength\tabcolsep{3pt} 
\begin{tabular} {l l 
l l l c l}
\toprule
Views & Modality & Studies & Methods & Highlights &  $\text{MPJPE}$ & Dataset\\
\midrule
\multirow{8}{*}{Monocular} &  \multirow{7}{*}{Vision}& Ching{-}Hang et al. \cite{Ching20173D} & CNN & A matching method to lift poses & 82.72 & \multirow{7}{*}{Human3.6m} \\
&  & Julieta et al. \cite{Martinez2017Simple} & CNN & Off-the-shelf detectors \& lifting networks & 87.3 & \\
& & Dushyant et al. \cite{singleshot}& CNN & Occlusion-robust pose-maps & 69.9 & \\
& & Zhao et al. \cite{zhao2019semantic} & CNN+GCN & An novel SemGCN for 2D-3D lifting. &  60.8 & \\
& & Shichao et al. \cite{li2020cascaded} & CNN  & \makecell[l]{Evolutionary 2D-3D data augmentation \\ \& Cascaded 2D-3D lifting networks} & 50.9 & \\
& & Kehong et al. \cite{gong2021poseaug} & CNN  &  Differentiable pose augmentor for 2D-3D lifting & 50.2 & \\
& & Yujun et al. $\cite{cai2019exploiting}^\dag$ & CNN+GCN & GCN-based 2D-3D sequence lifting. & 48.8 & \\
& & Wenbo et al. $\cite{hu2021conditional}^\dag$ & CNN+GCN & \makecell[l]{Graph for skeleton representation \\ \& GCN for 2D-3D sequence lifting} & 41.1 & \\
\cline{2-7}
& \makecell[l]{Vision \\ \& IMUs} & Timo et al. \cite{von2018recovering} & CNN & Video Inertial Poser to fusing images and IMUs & 26 & 3DPW\\
\midrule
\multirow{9}{*}{Multi-view} & \multirow{5}{*}{Vision} & Karim et al. \cite{Iskakov2019learnable} & CNN & An end-to-end DNN triangulation method &17.7 &\multirow{4}{*}{Human3.6m}\\
&  & Zhang et al. \cite{zhang2020adafuse}& CNN & An adaptive multi-view fusion method & 19.5  &\\
&  & Yihui et al. \cite{epipolarTrans}& CNN & An epipolar transformer & 26.9 & \\
&  & Haibo et al. \cite{crossviewfusion}& CNN & A cross-view fusion network& 31.17 & \\ \cline{7-7}
& & Size et al. \cite{wu2021graph} & CNN+GCN & \makecell[l]{Learnable association matching \\ \& graph-based 3D pose refinement} & 15.84 & CMU Panoptic \\ \cline{2-7}
& \multirow{5}{*}{\makecell[l]{Vision \\ \& IMUs}} & Trumble et al. \cite{trumble2017total} & CNN  & A two-stream network fuses video and IMUs & 87.3 & \multirow{3}{*}{Human3.6m}\\
& & Gilbert et al. \cite{gilbert2019fusing} & CNN & Incorporating \cite{trumble2017total} with enhanced 3D HPE & 71.9 &  \\
& & Zhang et al. \cite{zhang2020fusing}& CNN & An orientation regularized pictorial model & 24.6 & \\
\cline{7-7}
& & Malleson et al. \cite{malleson2017real} & CNN & Optimization based on IMUs and poses & 62 & \multirow{2}{*}{Total Capture} \\
& & Huang et al. \cite{huang2020deepfuse}& CNN & DeepFuse for vision-IMU data fusion& 28.9 & \\
\bottomrule
\end{tabular}
\caption{The state-of-the-art approaches for image-based 3D HPE. The $\text{MPJPE}$ scores that are defined in \autoref{para:mpjpe},  can be obtained by testing the methods on the Human3.6M dataset. The symbol $\dag$ denotes the methods of using image sequence.}
\label{tab:image-based3dhpe}
\end{table*}

\subsection{Monocular 3D HPE}

The monocular 3D HPE task is to predict the 3D positions of human joints for a monocular image, which is known to be an ill-posed problem. 
DNNs have shown remarkable performance on predicting the depth of monocular images \cite{fu2018deep} and the monocular 3D HPE tasks \cite{Martinez2017Simple,rhodin2018unsupervised}.
The monocular 3D HPE methods can be generally categorized into two types: the single-stage and the two-stage methods.
The basic difference between these two types is that two-stage methods entail using off-the-shelf 2D predictors with the 2D HPE datasets. 
The single-stage methods predict 3D poses from images directly. 
By contrast, two-stage methods estimate 2D poses and then lift the 2D poses to 3D poses. 
In this subsection, we summarize the state-of-the-art works of both methods for monocular 3D HPE.

\subsubsection{Single-stage Approaches} \hfill

As DNN-based methods can automatically build mappings from 2D images to 3D poses, the single-stage 3D HPE can be generally viewed as an extension of 2D HPE.
Early works applied the regression paradigm to estimate 3D human pose directly.
Li et al. \cite{li20143d} originally applied an end-to-end approach with the deep neural network for 3D HPE by the combination of a joint detector and a joint regressor. 
Luvizon et al. \cite{luvizon20182d} proposed a multi-task framework to perform multi-task learning, which predicts 3D poses directly.

Compared to the early studies, there are many current single-stage approaches that apply heatmap representation to 3D HPE.
For example, Pavlakos et al. \cite{pavlakos2017coarse} proposed a single-stage method that predicted 3D heatmaps in voxel space and proposed a coarse-to-fine prediction scheme to reduce the large 3D heatmap cost.
To reduce the computational cost of direct predicting 3D voxel heatmaps, Nibali et al. \cite{nibali20193d} predicted the marginal 2D heatmaps to produce 3D coordinates.

For the single-stage methods in multi-person 3D HPE, they can be similarly categorized into the top-down and bottom-up methods, recalling the \autoref{fig:MPPE_pipeline}. 
The top-down methods directly predict 3D poses of the detected person in each proposed bounding box. 
Bottom-up methods estimate the 3D pose by detecting all 3D joints and grouping all joints into person-specific sets. 
Typically, Fabbri et.al \cite{fabbri2020compressed} proposed a volumetric heatmap autoencoder to estimate 3D joint locations and employed a distance-based heuristic strategy to associate the body joints of a person.
In summary, the framework of the top-down and bottom-up methods in monocular multi-person 3D HPE is similar to the frameworks in \autoref{fig:MPPE_pipeline}, while the joint points are distributed in the 3D space.

\subsubsection{Two-stage Approaches}  \hfill

Although the single-stage methods are efficient, the two-stage methods generally yield better performance as they could further benefit from 2D information and large-scale 2D HPE datasets. 
Many studies utilize off-the-shelf 2D estimators to produce 2D poses and followed by a 2D-3D lifting method. 
For example, Martinez et al. \cite{Martinez2017Simple} used an off-the-shelf 2D estimator to produce 2D poses and then employed a simple lifting neural network to predict 3D poses on the basis of 2D poses.
Zheng et al. \cite{zheng20213d} utilized vision transformer to implement the 3D lifting from 2D pose sequence.
For two-stage methods, weak-supervised approaches are often used to leverage the 2D-to-3D constraint \cite{papandreou2017towards}. 
Pavllo et al. \cite{pavllo20193d} fuse 2D poses sequence from videos to predict accurate 3D poses that exploit unlabeled videos in a semi-supervised manner. 

In multi-person 3D HPE, the two-stage methods also employ the top-down and bottom-up paradigms yet incorporate the two-stage process. 
For the top-down methods, 
Rogez et al. \cite{rogez2019lcr} employed a region proposal network to propose bounding boxes for targeted persons and developed a pose proposal network to estimate the human poses in these bounding boxes. 
Zanfir et al. \cite{zanfir2018deep} performed bottom-up 2D MPPE and then recovered 3D poses from 2D poses by a 3D pose decoding module.

Many approaches conducted the two-stage approaches of graph neural networks to the lifting in 3D HPE.
For two-stage approaches, a GCN-based lifting network generally takes the graph from 2D poses as the input to produce 3D poses.
Zhao et al. \cite{zhao2019semantic} proposed a semantic graph convolutional network to regress 3D joint coordinates from the 2D joint coordinates. 
Hu et al. \cite{hu2021conditional} proposed a directed graph-based skeleton representation and applied a graph convolutional network to exploit both spatial and temporal information of image sequences. 
The two-stage GCN-based HPE approaches show competitive performance by considering the graph of the prior knowledge of the human skeleton. 
However, GCNs-based approaches generally integrate other DNNs-based models to extract 2D poses in the first stage as the input of GCNs (the second stage), which limits the usage of GCNs in the HPE field.


\subsection{Multi-view 3D HPE}

The multi-view 3D HPE tasks aim to predict 3D joints position with synchronous multi-view cameras. 
Although the classic triangulation method (the details refer to \cite{multiviewgeometry}) can calculate accurate 3D object locations via multi-view camera systems, the triangulation method performs a drawback of noticeable sensitivity to inaccurate 2D prediction. 
Alleviating the error of 2D prediction is a critical problem of multi-view 3D HPE. 
Iskakov et al. \cite{Iskakov2019learnable} proposed an end-to-end DNN-based learnable triangulation method to produce confidence weights for each view. 
Qiu et al. \cite{crossviewfusion} proposed a multi-view fusion layer to improve 2D pose estimation and used a recursive pictorial structure model to predict 3D pose. 
He et al. \cite{epipolarTrans} introduced epipolar geometry to multi-view fusion, which significantly decreased the number of parameters in the fusion module. 

Similar to the multi-person HPE, the existing multi-view 3D HPE can be categorized into two types: the top-down and bottom-up methods. 
The top-down methods predict the 2D poses of each view and match the 2D poses to perform the 3D reconstruction.
Dong et al. \cite{dong2019fast} proposed a multi-way matching algorithm that combined both geometric and appearance cues to match the detected 2D poses across views. 
Chen et al. \cite{chen2020cross} utilized the temporal consistency to match multi-view 2D poses to 3D poses, and retained and updated the 3D poses by the cross-view multi-person tracking. 
Huang et al. \cite{huang2020end} proposed a dynamic matching algorithm to match corresponding multi-view 2D poses from different views for each person and then used point triangulation to recover 3D poses.
Currently, there are a few studies that apply bottom-up approaches to multi-view 3D HPE.
Elmi et al. \cite{elmi2021light3dpose} originally applied a bottom-up approach to the multi-view 3D HPE. 
The 2D features of each image were processed by a backbone network and then aggregated by an un-projection layer into a 3D input representation. 
Finally, a sub-voxel joint detection module and a skeleton decoder module were employed to produce a set of 3D poses.

\subsection{Multimodal Learning in 3D HPE}

Multimodal learning is a deep learning-based approach that builds models with multiple modalities. 
A modality often refers to a sensory modality such as vision, touch signal, or radio signal. 
With the fusion of multiple sensors, multimodal learning approaches are more robust and capable of overcoming the challenges like occlusions than vision methods \cite{zhao2018through}.
Although researchers have shown increasing interest in the field of multimodal learning approaches for HPE, a few studies apply multimodal learning to implement 3D HPE. 
We retrieve the related studies of multimodel learning in the Scopus database, as shown in \autoref{fig:multimodal_learning_hpe}. 


In this subsection, we review the related deep learning-based approaches that utilize 2D vision data, depth information, and multimodal information (e.g, IMUs signal).
Marin et al. \cite{marin20183d} proposed a deep depth pose model to combine RGB-D information and a set of predefined 3D poses to predict the 3D joint positions. 

Furthermore, the deep learning-based methods with the data of IMUs and images have shown remarkable results in 3D HPE (as shown in \autoref{tab:image-based3dhpe}). 
Marcard et al. \cite{von2018recovering} adopted IMUs and a mobile camera to estimate 3D human poses, and proposed the skinned multi-person linear model to produce initial 3D poses and then associated the 3D poses with 2D detected persons. 
Huang et al. \cite{huang2020deepfuse} used the orientation of body parts, that is captured by IMUs, to refine the image-based pose estimation. 
In summary, multimodal HPE generally shows superior performance compared to vision-only HPE.
However, it’s worth noting that the lack of labeled datasets is a critical problem in multimodal HPE.

\section{Applications}
\label{sec:applications}

HPE has been used in a variety of applications such as action analysis \cite{liu2020disentangling}, HCI \cite{shotton2011real}, gaming \cite{ke2010real}, sport analysis \cite{wang2019ai,park2017accurate}, motion capture \cite{yang2020transmomo}, computer-generated imagery \cite{hornung2007character,willett2020pose2pose}.
Accurate joint points are semantically informative and can be utilized in computer vision tasks like action recognition, and human tracking \cite{luvizon20182d}.
In this section, we present the main applications of HPE (see 
the summary in \autoref{tab:applications}).
\noindent
\begin{table*}[!ht]
\centering \footnotesize
\renewcommand{\arraystretch}{1.1}
\setlength\tabcolsep{4pt} 
\begin{tabular} {l l l m{5.9cm}}
\toprule
2D/3D HPE & Applications & Methods  & Remarks \\
\midrule
\multirow{4}{*}{2D HPE} & \makecell[l]{Action \\ Analysis} & \makecell[l]{$\bullet$ Bottom-up MPPE (OpenPose \cite{cao2017realtime}) + GCN \cite{kim2019skeleton} \\ $\bullet$ Top-down MPPE (AlphaPose \cite{fang2017rmpe}) + GCN}  & HPE provides spatial joint data and implicit temporal correlationship for action analysis.
\\ \cline{2-4}
& \makecell[l]{Character \\ Animation} & \makecell[l]{$\bullet$ Top-down MPPE \cite{wei2016convolutional} + Motion Transfer \cite{weng2019photo} \\ $\bullet$ Bottom-up MPPE \cite{cao2017realtime} + Motion Transfer \cite{willett2020pose2pose}} & Animation by HPE-based motion transferring from human performer to animated character \\
\cline{2-4}
& \makecell[l]{Sports \\ Analysis} & \makecell[l]{ $\bullet$ Bottom-up MPPE \cite{cao2017realtime} + Coaching method \cite{su2018position}\\
$\bullet$ Video-based HPE + Coaching method \cite{wang2019ai}} & HPE-based sports analysis by comparisons between players and exemplars \\
\cline{2-4}
& \makecell[l]{Medical \& \\ Clinical} & \makecell[l]{$\bullet$ Top-down MPPE + Exercise Supervision \cite{li2020human} \\ $\bullet$ Bottom-up MPPE + Exercise Supervision \cite{rabbito2021using}}  & Joint positions provide rich information for clinical applications like in-bed/sleep monitoring \\
\midrule
\multirow{5}{*}{3D HPE}  & \makecell[l]{Action \\ Analysis} & \makecell[l]{ $\bullet$ Multimodal 3D HPE (Kinect-based) + Random Forest \cite{hbali2017skeleton}\\
$\bullet$ Multimodal 3D HPE (Kinect-based) + CNN \cite{tsai2019implementation}} & 3D action analysis can exploit both 2D joint position and extra depth information \\
\cline{2-4}
& \makecell[l]{HCI} & \makecell[l]{$\bullet$ Multimodal 3D HPE (Kinect-based) + HCI \cite{shotton2011real}\\
$\bullet$ Monocular 3D HPE + HCI \cite{ke2010real}
}  & HPE-based HCI is a natural and contactless way \\
\cline{2-4}
& \makecell[l]{Character \\ Animation} & \makecell[l]{$\bullet$ Two-stage Monocular 3D HPE + Motion Transfer \cite{kumarapu2021animepose,yang2020transmomo}} & 3D character animations benefit from 3D HPE and motion transfer \\
\cline{2-4}
& \makecell[l]{Sports \\ Analysis} & \makecell[l]{$\bullet$ Multimodal 3D HPE (Kinect-based) + Coaching Method \cite{elaoud2020skeleton,park2017accurate} \\ $\bullet$ Monocular 3D HPE \cite{li20143d} + Coaching Method \cite{kamel2019investigation}} & 
3D HPE provides more dimensionality (3D angles between joints) for a coaching system \\
\cline{2-4}
& \makecell[l]{Medical \& \\ Clinical} & \makecell[l]{$\bullet$ Multimodal 3D HPE (Kinect-based) + Neural Recording \cite{gabriel2016neural}\\
$\bullet$ Multimodal 3D HPE (Kinect-based) + Exercise Supervision \cite{obdrvzalek2012accuracy}} & 3D HPE-based clinical monitoring and rehabilitation system use human joint and depth information \\
\bottomrule
\end{tabular}
\caption{Summary of the approaches in HPE-based applications.}
\label{tab:applications}
\end{table*}


\subsection{Action Analysis}
Action recognition/prediction is mainly a temporal task based on image sequences. 
The traditional methods might demand extensive computation and appear unstable to environmental variations in terms of illuminations, objects in the background or foreground, body scale, and motion blur \cite{ren2020survey}. 
The human skeleton is naturally a high-level representation which has shown benefits for action analysis tasks such as action recognition \cite{liu2019ntu} and action detection \cite{li2017skeleton}. 
For example, Duan et al. \cite{duan2022revisiting} stacked 2D heatmaps from human pose sequences to 3D heatmap volumes and utilized ResNet layers to predict human actions from these volumes. 
Liu et al. \cite{liu2020disentangling} used graph convolutional network to utilize 3D pose sequence for action analysis.

Currently, with the development of sensors (e.g., Kinect \cite{obdrvzalek2012accuracy}) and HPE algorithms, 
large-scale and accurate skeleton data for action recognition \cite{liu2019ntu} becomes accessible.
The space correlations and temporality of skeleton sequences provide informative prior knowledge to yield a robust motion pattern. 
On large-scale action recognition dataset NTURGB-D \cite{liu2019ntu}, the state-of-the-art skeleton-based methods \cite{duan2022revisiting,das2020vpn} achieved more than 95\% accuracy, while image-only methods \cite{luvizon20182d} achieved less than 90\% accuracy. 



In addition, the skeleton-based action analysis can be used to build smart surveillance systems. 
Hbali et al. \cite{hbali2017skeleton} used HPE and action analysis to build up an elderly monitoring system for alerting dangerous activities. 
Guo et al. \cite{guo2018image} used skeleton-based action recognition to identify unsafe behaviours of construction workers. 
In conclusion, HPE provides significant spatial joint information and implicit temporal correlations for skeleton-based action analysis.


\subsection{Human-Computer Interaction}
Human-computer interaction (HCI) has been studied for several decades and plays an important role in our daily lives. 
Traditional HCI techniques allow humans to interact with computers via tangible devices or interfaces such as mice, keyboards, or touch screens. 
Compared with traditional interactions, HPE-based HCI provides a natural and contactless way that is highly suitable for the difficult pandemic situation of COVID-19, particularly when using public devices.

HPE-based HCI techniques are widely applied in many applications such as arts, gaming, and virtual reality. 
Most vision-only applications consider the estimation of 2D human poses since currently, 2D HPE approaches can offer more accurate and prompt predictions than 3D HPE. 
While 3D HPE often employs depth-aware sensors (like RGB-D cameras) to produce reliable and informative 3D poses.
The well-known depth-aware distributed Microsoft Kinects \cite{obdrvzalek2012accuracy} is designed to capture human body movements for gaming and virtual reality teleconference systems \cite{lan2016development}. 
Kamel et al. \cite{kamel2019investigation}, Thar et al. \cite{thar2019proposal}, and Park \cite{park2017accurate} et al. used 3D monocular HPE and a single camera to capture human poses for producing action evaluation and feedback for Tai Chi, Yoga, and golf, respectively.  
In summary, HPE-based HCI is a type of natural and contactless interface that differs from the traditional HCI.

\subsection{Character Animation} 
Creating high-quality character animation is important in animation films and computer games. 
Traditional methods rely on creating costly and time-consuming frame-wise animations. 
HPE-based performance-driven animations accomplish character motions by transferring motions from a human performer to an animated character, which is prevalent in the film and game industries conveniently and economically.
For 2D animation applications, Willett et al. \cite{willett2020pose2pose} proposed a novel method to jointly yield 2D animation and 2D character creation by leveraging human pose.
In addition, the HPE technique is helpful in augmented reality. 
For instance, Weng et al. \cite{weng2019photo} utilized human skeleton information via HPE to generate 3D character animation from an image. 
To generate 3D animations, a commonly used method is to control all parts of a character via motion capture \cite{yang2020transmomo}. 
In summary, DL-based HPE approaches provide an alternative to traditional motion capture for creating both 2D and 3D animations.


\subsection{Sports Analysis}
Sports performance analysis provides statistics/recording for coaches and players to improve their performance. 
The automatic sports analysis demands precise postures and motion of players, which requires accurate markerless motion capture techniques. 
The HPE approaches perform effectively in the analysis of many sporting activities.
HPE-based methods can be used to compare the differences between learners and instructors in various sports. 
For example, Park et al. \cite{park2017accurate} used the HPE approach to analyze activities in golf. 
By comparing a user's swing with the reference swing, the user can examine his/her evaluation result in head movement, knee alignment, swing rhythm, and balance of golf swing in their approach. 
Kamel et al. \cite{kamel2019investigation}, and Thar et al. \cite{thar2019proposal} employed HPE to evaluate the difference between a learner's posture and an instructor's posture in Tai Chi and Yoga, respectively. 
Wang et al. \cite{wang2019ai} built up an HPE-based athletic training assistance system to detect bad poses from a sequence of 2D users' poses.
In addition, HPE-based approach is applied to capture players' motion in sports like badminton \cite{su2018position}, soccer \cite{afrouzian2016pose,bridgeman2019multi}, and tennis \cite{kurose2018player,wu2020futurepong}. 
In conclusion, HPE-based sports analysis mainly relies on a comparison between a player and an exemplar.
2D HPE is more commonly adopted than 3D HPE as 3D HPE currently requires either multi-view setups or depth-aware sensors.

\subsection{Medical and Clinical Applications}
Accurate joint positions provide rich information for clinical applications like in-bed monitoring, sleep laboratories, epilepsy monitoring, and intensive care units.
Gabeiel et al. \cite{gabriel2016neural} combined synchronized Kinect v2 and standard clinical electrocorticography monitors, which record neural activity from the cortex, to investigate the relationship between human movement behaviours and neural activity. 
In rehabilitation medicine, HPE is successfully applied to improve the rehabilitation of patients.
Obdr{\v{z}}{\'a}lek et al. \cite{obdrvzalek2012accuracy} employed HPE for observation and online feedback when coaching elderly patients with the daily physical exercise routine. 
Li et al. \cite{li2020human} introduced an in-home lower-limb rehabilitation system with an HPE-based model to help patients perform rehabilitation activities even without the presence of physical therapists. 
In \cite{rabbito2021using}, Rabbito used OpenPose \cite{cao2017realtime} and motion capture system Vicon \footnote{\url{https://www.vicon.com/}} to analyze patient gait for rehabilitation medicine.

In summary, HPE-based clinical applications generally fall into two main types, in-bed monitoring and rehabilitation training.
Clinical environments are suitable for installing multi-view cameras or depth-aware sensors, which encourages 3D HPE. 
HPE-based Rehabilitation training currently is similar to HPE-based sports coaching, which is mostly based on the comparison between a user and an exemplar.
While traditional motion capture systems are available in limited environments, HPE-based methods have great potential for providing more extensive human motion data.

\section{Research Trends and Challenges}
\label{sec:trends_chall}
In this section, we aim to point out the observed research trends via bibliometrics, raise the challenging issues, and provide insightful recommendations for future research.

\subsection{Research Trends}
\label{subsec:trends}

In this subsection, we identify the research trends via bibliometrics which refers to the use of statistical methods to analyze books, articles and other publications. 
It is an effective way to measure publications in the scientific community.
However, current survey papers rarely use bibliometrics to analyze the data of publications, particularly in HPE. 
This paper applies bibliometrics to retrieve the publications for discovering and demonstrating the research trends in HPE. 

\begin{figure*}[!ht]
\centering
    \begin{subfigure}[t]{.25\textwidth}
        \centering
        \includegraphics[width=\textwidth]{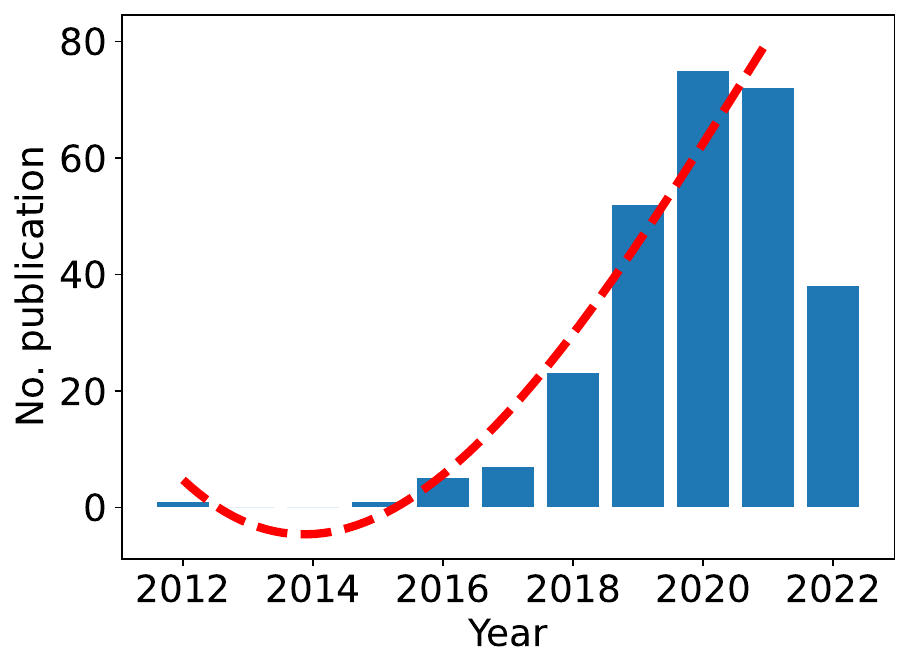}
        \caption{}
        \label{fig:multi_person}
    \end{subfigure} \hspace{-2mm}
    \begin{subfigure}[t]{.24\textwidth}
        \centering
        \includegraphics[width=\textwidth]{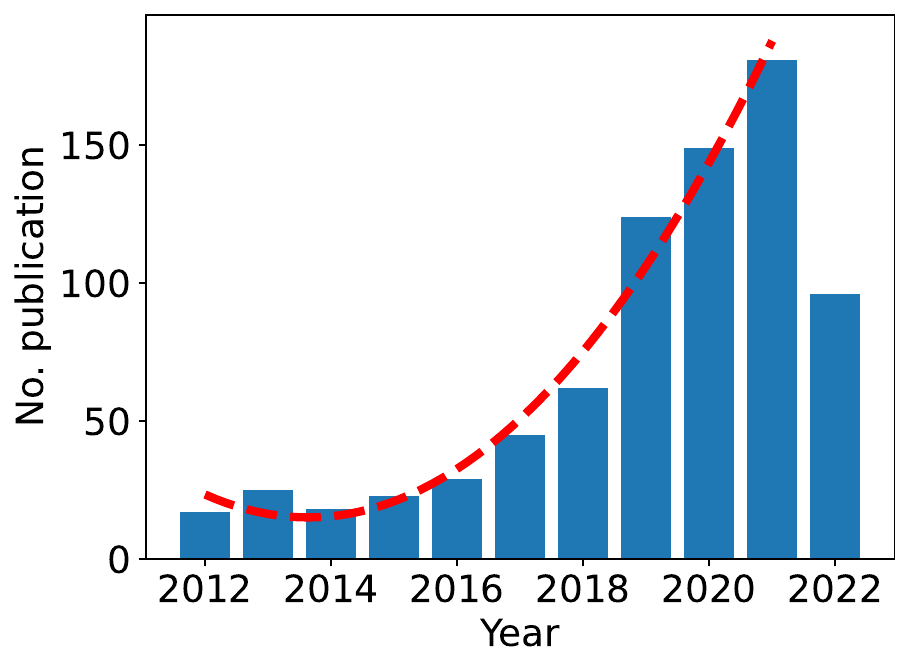}
        \caption{}
        \label{fig:3d_hpe}
    \end{subfigure} \hspace{-2mm}
    \begin{subfigure}[t]{.24\textwidth}
        \centering
        \includegraphics[width=\textwidth]{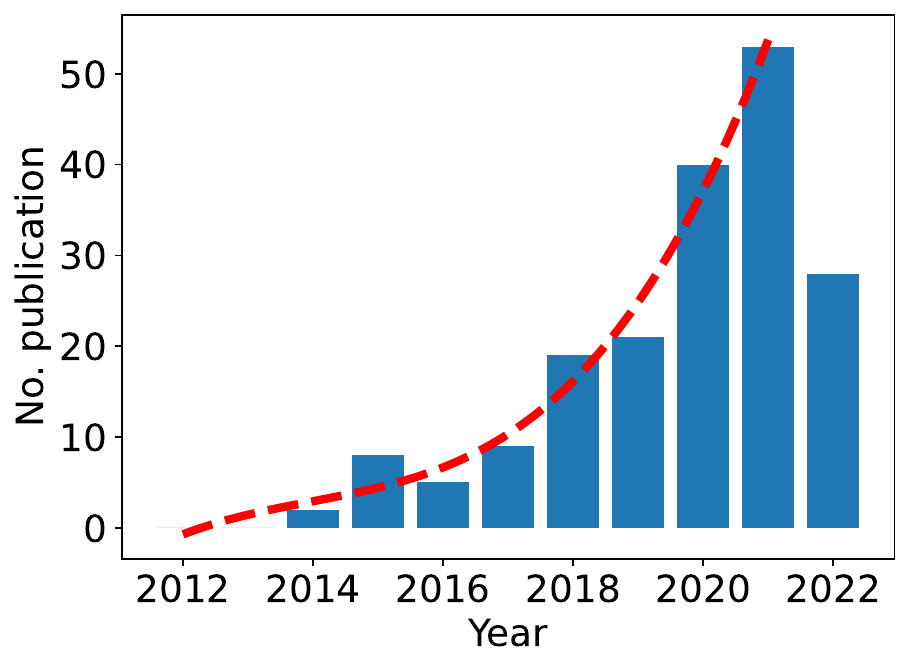}
        \caption{}
        \label{fig:efficient_hpe}
    \end{subfigure} \hspace{-2mm}
    \begin{subfigure}[t]{.24\textwidth}
        \centering
        \includegraphics[width=\textwidth]{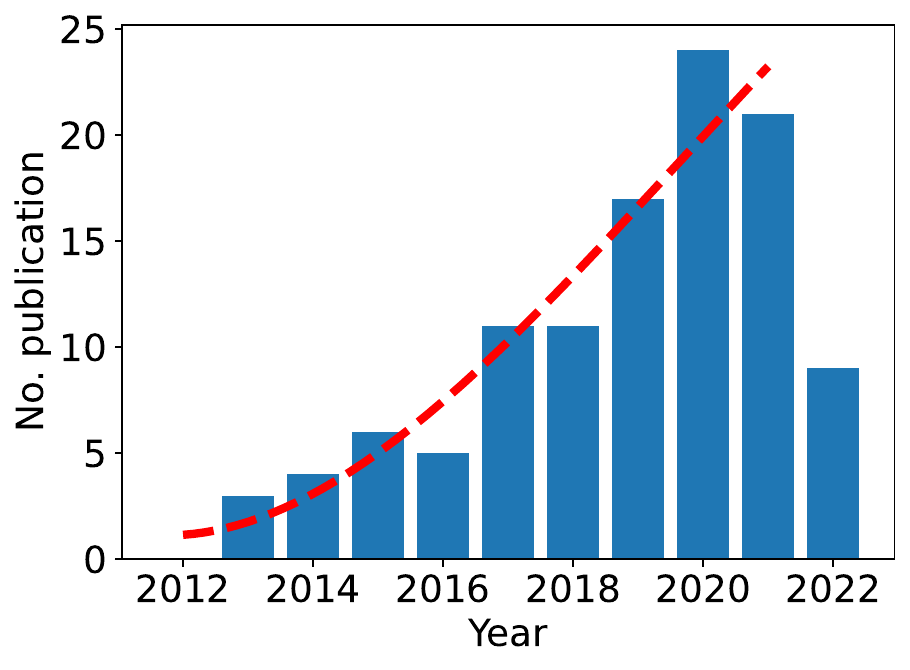}
        \caption{}
        \label{fig:multimodal_learning_hpe}
    \end{subfigure}
    \caption{
    The number of publications on the HPE topics of a) multi-person pose estimation, b) 3D HPE, c) efficient deep learning-based HPE, d) multimodal learning HPE from Scopus. Scopus database returned 274, 769, 185, 111 results for a), b), c), and d) respectively until 08/09/2022. To observe the trends clearly, we retrieve the data of each year over the ten years from 2012 to 2022. The red dashed lines are the fit of the number of publications from 2012 to 2021 (NOT 2022). The source code to retrieve the data are available at \url{https://github.com/wuyuuu/elsevier-search}.
    }
    \label{fig:open_source}
\end{figure*}

In general, bibliometric data can be obtained from various databases.
In this paper, we choose the Scopus database \footnote{\url{https://en.wikipedia.org/wiki/Scopus}} for literature retrieval.
Scopus is the largest abstract/citation database with peer-reviewed literature, which is released by Elsevier. 
The resources of Scopus are more accurate and comprehensive than other alternatives such as PubMed, Web of Science, and Google Scholar \cite{martin2018google}.
Importantly, Elsevier provides a Python library \footnote{\url{https://github.com/pybliometrics-dev/pybliometrics}} to retrieve the data for the expected topics from Scopus database \cite{rose2019pybliometrics}.
In this work, we review the articles in vision-based HPE using deep learning by retrieving "title, abstract, and keywords" in the Scopus database.
We present the trends that are observed from the literature in the four aspects: multi-person pose estimation, 3D HPE, efficient deep learning-based HPE, and multi-modal learning.
To observe the trends clearly, we retrieve the data for each year over ten years (2012 - 2022).
We retrieve the number of publications in Scopus data from the four aspects, as shown in \autoref{fig:open_source}, where the red dash lines are the fit of the number of publications from 2012 to 2021 (NOT 2022).

\subsubsection{Multi-Person HPE}\hfill

Multi-Person HPE has become an important research topic.
We retrieve the literature on this topic by using the search code TITLE-ABS-KEY(("multi-person" or "crowd") and ("multi-person pose estimation" or "human pose estimation")). 
The number of publications over the years (from 2012 to 2022) is shown in \autoref{fig:multi_person}. 
It shows a very significant increasing trend in the number of Multi-Person HPE publications since 2015, while an insignificant drop in 2021.

On one hand, the small number of publications shows that the multi-person pose estimation in crowd scenes is still rarely investigated. 
On the other hand, the increasing number of publications shows the increasing trend in this topic. 
This technique has been applied in various applications such as multi-person gaming and sports analysis.
However, as reported in \cite{li2019crowdpose}, the performance of the existing methods degrades dramatically as the crowd level increases.
In multi-person HPE, there are still many open issues and challenges that need to be further studied in the future. 

\subsubsection{3D HPE} \hfill

Accurately estimating 3D positions for human joints is an important topic in computer vision.
Compared to 2D HPE, 3D HPE could provide extra depth information which brings broader applications such as markerless motion capture, video games, and sports analysis \cite{park2017accurate,trumble2017total}.
We observe that 3D HPE attracts more and more research interest recently.
We design the code of TITLE-ABS-KEY(("3D") and ("human pose estimation")) to retrieve the literature on the 3D HPE-related topics in the Scopus database.
The number of publications over the years (from 2012 to 2022) is shown in \autoref{fig:3d_hpe}. 

We observe an increasing number of 3D HPE publications from 2014 to 2021.
However, the existing 3D HPE methods still need to be investigated for desired performance because of unsolved problems such as the ill-posed property, the lack of outdoor environment annotations, and the incapability of performing real-time inference on edge devices. 
Therefore, the 3D HPE will be a crucial research trend in the future.

\subsubsection{Efficient Deep Learning-based HPE} \hfill

To apply HPE methods to real applications, the efficiency of HPE approaches is a crucial research topic. 
In particular, although deep learning-based methods have achieved state-of-the-art performance in HPE, the approaches of deep neural networks may suffer from high computational costs and inference delay.
We design the code of TITLE-ABS-KEY(("efficient" or "real-time") and ("human pose estimation")) to retrieve literature in efficient HPE. 
The number of publications over year (from 2012 to 2022) is shown in \autoref{fig:efficient_hpe}.
It shows a significant increase since 2014 with respect to the number of publications in efficient HPE.
We note that the HPE efficiency problem was generally tackled by designing efficient networks manually in previous approaches \cite{rafi2016efficient,cao2017realtime}, while current studies \cite{mcnally2021evopose2d,xu2021vipnas} introduce neural architecture search (NAS) into HPE, which might be a typical method in solving the efficiency problem of neural network in the future.

\subsubsection{Multimodal Learning for HPE} \hfill

Multimodal learning extracts features from multiple sensing modalities, which alleviates the complexity of unimodal methods. 
Recently, the multimodal learning-based approaches \cite{zhang2020fusing,von2018recovering} have been validated for robustness and accuracy in HPE tasks. 
We retrieve the literature by using the code of TITLE-ABS-KEY(("multi-modal" or"multimodal" or "IMUs" or "radio signal") and ("human pose estimation") and not ("distribution")) to observe and demonstrate the research trend. 
The number of publications over years (from 2012 to 2022) is shown in \autoref{fig:multimodal_learning_hpe}. 
Although the total number of publications implies a lack of attention in multimodal learning HPE, the steady growth of publications implies an increasing interest in this topic.
The multimodal learning methods offer solutions and robustness to the vision-based model in addressing occlusions, which are promising solutions towards the in-the-wild challenge \cite{von2018recovering}. 
The multimodal learning methods for HPE still need to be further investigated in the future.

\subsection{Research Challenges}
\label{subsec:challeges}

A remarkable HPE approach should consider both high accuracy and high efficiency. 
Although many studies have investigated the HPE with prominent performance, there are many considerable challenges to achieve both goals.
An accurate HPE approach is generally demanded to deal with the challenges of various occlusions and personal appearances, estimate the depth of monocular image, and remain robust to image degeneration. 
In addition, the current biased datasets (e.g., incomprehensive outdoor 3D annotations, relatively rare uncommon poses) is a prevalent challenge for the practical accuracy of HPE in real-world scenarios.
Moreover, implementing efficient deep learning-based HPE on resource-limited devices is a notorious challenge. 
The current deep learning-based HPE generally take a lot of computing time because of the large-scale networks.
In particular, the computing time significantly creases with the creasing number of people for the crowd scenes. 
Therefore, a number-robust HPE algorithm is critical for implementing an efficient HPE.
In this subsection, we summarize and discuss the challenges in HPE from two aspects: accuracy and efficiency.

\subsubsection{Challenges in Accurate HPE} \hfill 

\paragraph{Diverse Human Poses and Appearances} 
A fundamental challenge in developing accurate HPE comes from the diversity of human poses \cite{moeslund2001survey}. 
Despite the vast range of human appearances, the human body entails high degree of freedom, demanding an advanced presentation ability for data-driven approaches in HPE. 
Additionally, image degeneration like motion blur and image defocus exists in video-based data \cite{luo2018lstm}, which also hinders HPE approaches from achieving remarkable performance.

\paragraph{Occlusions} 
Although current HPE methods perform outstanding performance on many public datasets, a well-known issue is noticeable performance degeneration caused by occlusions and highly deformable human body \cite{andriluka14cvprmpii,fang2017rmpe,cheng2020bottom}.
Self-occlusions and mutual occlusions could prompt the occlusions and environmental truncation, while mutual occlusions can occur extensively in crowd scenarios \cite{li2019crowdpose} to cause the performance to decline dramatically. 
The highly deformable human body can cause ambiguity in small-scale human instances \cite{cheng2020bottom} or the specific human poses as well. 
Thus, designing the powerful HPE method for occlusion scenarios with the capability of utilizing global information and prior knowledge would be a challenging issue.

\paragraph{Incomprehensive Datasets}
To apply HPE approaches to practical applications, challenges could come from the gap between current incomprehensive datasets and real-world applications. 
For example, uncommon poses like falling down are less likely to appear in datasets, and the outdoors 3D HPE dataset is relatively rare. 
This gap leads to an imbalanced learning problem, which could hinder the applications of 3D HPE in the real world. 
Although current approaches can leverage semi-supervised learning \cite{pavllo20193d} or synthetic dataset \cite{fabbri2018learning} to enrich the datasets, the semi-supervised methods still need a lot of quality training data \cite{rhodin2018unsupervised}.
Current datasets lack realistic simulation of lighting effects, clothing meshes, and environment interactions \cite{doersch2019sim2real}. 
Therefore, it is difficult to transfer the trained neural networks from the simulation to real-world applications. 
Developing a remarkable HPE on current incomprehensive datasets is still challenging, particularly for deployment in complex environments.

\begin{table*}[!ht]
\centering
\footnotesize
\renewcommand{\arraystretch}{1.1}
\setlength\tabcolsep{2pt}
\begin{tabular}{l l m{7.5cm} m{5.1cm}}
\toprule
Types & Studies & URL (Open source code) & Remarks\\ 
\midrule
2D/3D HPE & MMPose & {\footnotesize \url{https://github.com/open-mmlab/mmpose}} & Well-known platform \\
\midrule
\multirowcell{13}{2D\\ HPE}
& Associative embedding \cite{newell2017associative} & {\footnotesize \url{ https://github.com/princeton-vl/pose-ae-train}} & A SOTA grouping for bottom-up approaches \\
& Hourglass \cite{newell2016stacked} & {\footnotesize \url{ https://github.com/princeton-vl/pose-hg-demo}} & Effective yet simple backbone \\
& OpenPose \cite{cao2017realtime}& {\footnotesize \url{https://github.com/CMU-Perceptual-Computing-Lab/openpose}} & Real-time \& bottom-up \\
& AlphaPose \cite{fang2017rmpe}& {\footnotesize \url{https://github.com/MVIG-SJTU/AlphaPose}} & Real-time \& top-down\\
& Higher-HRNet \cite{cheng2020higherhrnet}& {\footnotesize \url{https://github.com/HRNet/HigherHRNet-Human-Pose-Estimation}} & A SOTA bottom-up approach\\
& HRNet \cite{sun2019deep}& {\footnotesize \url{https://github.com/HRNet/HRNet-Human-Pose-Estimation}} & A SOTA top-down approach\\
& RLE \cite{li2021human} & {\footnotesize \url{https://github.com/Jeff-sjtu/res-loglikelihood-regression}} & A SOTA regression-based HPE \\
& UDP-POSE \cite{huang2020devil} & {\footnotesize \url{https://github.com/HuangJunJie2017/UDP-Pose}} & 1st in ICCV 2019 COCO keypoint challenge \\
& DARK \cite{zhang2020distribution} & {\footnotesize \url{https://ilovepose.github.io/coco/}} & 2nd in ICCV 2019 COCO keypoint challenge \\
& Lite-HRNet \cite{yu2021lite}& {\footnotesize \url{https://github.com/HRNet/Lite-HRNet}} & Lightweight HRNet-based model \\
& Lightweight OpenPose \cite{osokin2018real}& {\footnotesize \url{https://github.com/Daniil-Osokin/lightweight-human-pose-estimation.pytorch}} & Real-time on CPU \\
& BlazePose \cite{bazarevsky2020blazepose} & {\footnotesize \url{https://google.github.io/mediapipe/solutions/pose.html}} & Real-time 2D single-person HPE driven by MediaPipe \cite{lugaresi2019mediapipe} \\
& PRTR \cite{li2021poseTransformer} & {\footnotesize \url{https://github.com/mlpc-ucsd/PRTR}} & 2D Pose 
Regression transformer \\
& DCPose \cite{liu2021deep} & {\footnotesize \url{https://github.com/Pose-Group/DCPose}} & 1st in PoseTrack2017 \& PoseTrack2018 \\
\midrule
\multirowcell{12}{3D\\ HPE}
& Epipolar transformer \cite{epipolarTrans} & {\footnotesize \url{https://github.com/yihui-he/epipolar-transformers}} & A SOTA multi-view approach \\
& Learnable triangulation \cite{Iskakov2019learnable} & {\footnotesize \url{https://saic-violet.github.io/learnable-triangulation/}} & A SOTA multi-view approach \\ 
& SMAP \cite{zhen2020smap} & {\footnotesize \url{https://github.com/zju3dv/SMAP}} & SOTA single-view multi-person  \\
& DOPE \cite{weinzaepfel2020dope} & {\footnotesize \url{https://github.com/naver/dope}} & Real-time whole-body 3D HPE \\
& VNect \cite{mehta2017vnect}& {\footnotesize \url{http://gvv.mpi-inf.mpg.de/projects/VNect/}} & Real-time single-view approach \\
& Synthetic occlusion \cite{sarandi2018synthetic} & {\footnotesize \url{https://github.com/isarandi/synthetic-occlusion}} & 1st place in ECCV2018 3D HPE Challenge \\
& Integral regression \cite{sun2018integral}
& {\footnotesize \url{https://github.com/JimmySuen/integral-human-pose}} & 2nd place in ECCV2018 3D HPE Challenge\\
& PoseFormer \cite{zheng20213d} & {\footnotesize \url{https://github.com/zczcwh/PoseFormer}} & 3D pose transformer \\
& \makecell[l]{Top-down \& bottom-up \\ Integration \cite{cheng2021monocular}} & {\footnotesize \url{https://github.com/3dpose/3D-Multi-Person-Pose}} & A SOTA monocular multi-person 3D HPE \\
& Normalizing flows \cite{wehrbein2021probabilistic} & {\footnotesize \url{https://github.com/twehrbein/Probabilistic-Monocular-3D-Human-Pose-Estimation-with-Normalizing-Flows}} & A SOTA monocular 3D HPE \\
& PoseAug \cite{gong2021poseaug} & {\footnotesize \url{https://github.com/jfzhang95/PoseAug}} & A data augmentation framework for 3D HPE\\
\bottomrule
\end{tabular}
\caption{The open-source code of the state-of-the-art HPE methods.}
\label{tab:open-source}
\end{table*}

\subsubsection{Challenges in Efficient HPE} \hfill
\paragraph{Computation-intensive Neural Networks} \hfill

Eventually, the HPE approach needs to be implemented for applications in the real world. 
However, the state-of-the-art neural networks \cite{sun2019deep,zhen2020smap} are generally hard to be implemented on mobile devices or embedded devices as their enormous computational cost. 
Thus, it is crucial to design lightweight neural networks for efficient HPE. 
The existing methods of designing lightweight neural networks are mainly manual design and heuristic design (e.g., NAS \cite{xu2021vipnas}).
However, the method of manual design is hard to balance the accuracy and network size \cite{yu2021lite}.
NAS-based methods \cite{xu2021vipnas,mcnally2021evopose2d} generally need various computational cost even weeks of CPU time. 
Therefore, developing lightweight neural networks for HPE is still a challenging task.

\paragraph{Time-Consuming MPPE} \hfill

Currently, MPPE algorithms consume increasing computation time over the increasing number of targeted persons.
Top-down approaches estimate the pose of each detected person after the person detection stage.
Bottom-up approaches \cite{newell2017associative,cao2017realtime} predict similarity values between keypoints and employ a matching algorithm (e.g., Hungarian algorithm) for grouping keypoints. 
Note that top-down and bottom-up approaches are two-stage methods due to top-down approaches require an extra detection stage and bottom-up approaches require an extra grouping stage, besides estimating keypoint locations. 

Compared to two-stage methods, single-stage methods generally perform superior in terms of computational cost. 
A promising single-stage method \cite{nie2019single} predicts the locations of persons and keypoints' offsets to each location.
However, the single-stage methods are generally not as competitive as the state-of-the-art two-stage methods \cite{cheng2020higherhrnet,sun2019deep} in terms of accuracy. 
How to develop a desired efficient MPPE is still an interesting challenge, particularly for applications in the real world.

\section{Conclusions}
\label{sec:conclusion}

In this paper, we presented an up-to-date and in-depth overview of the deep learning approaches in vision-based HPE.
We systematically introduced the preliminary knowledge in HPE and reviewed the HPE approaches in two categories: 2D-based approaches and 3D-based approaches. 
We discussed a number of interesting applications of deep learning-based HPE. 
Finally, we pointed out the research trends via bibliometrics, raised the challenging issues, and provided insightful recommendations for future research.
To help readers to reproduce the state-of-the-art methods, we summarized the open-source codes of the well-known studies for 2D and 3D deep learning-based HPE in \autoref{tab:open-source}, which could help readers easily implement their HPE tasks. 
This article provides a meaningful overview as introductory material for beginners to deep learning-based HPE, as well as supplementary material for advanced researchers.

\bibliographystyle{IEEEtran} 
\bibliography{main}

\end{document}